# Log anomaly detection via Meta Learning and Prototypical Networks for Cross domain generalization

Krishna Sharma[1], Vivek Yelleti[1]

[1]Department of Computer Science and Engineering, SRM University AP, India

**Abstract.** Log anomaly detection is essential for system reliability, but it is extremely challenging to do considering it involves class imbalance. Additionally, the models trained in one domain are not applicable to other domains, necessitating the need for cross-domain adaptation (such as HDFS and Linux). Traditional detection models often fail to generalize due to significant data drift and the inherent absence of labeled anomalies in new target domains. To handle the above challenges, we proposed a new end-to-end framework based on a meta-learning approach. Our methodology first gets the data ready by combining a Drain3 log parsing mechanism with a dynamic drift-based labeling technique that uses semantic and fuzzy matching to move existing anomaly knowledge from one source to another. BERT-based semantic embeddings are obtained, and the feature selection is invoked to reduce the dimensionality. Later, Model Agnostic Meta-Learning (MAML) and Prototypical Networks models are trained to adapt quickly and effectively. The SMOTE oversampling method is employed to handle imbalances in the data. All the results are obtained by employing the leave-one-out source method, and the corresponding mean F1 scores are reported. Our empirical findings validate that the proposed meta-learning-driven approach yielded the highest mean F1 score and proved to be effective for cross-domain settings.

**Keywords:** Anomaly detection; Meta Learning; BERT; Log Mining

## 1  Introduction

System logs are essential, typically voluminous, and is generally textual records of runtime events, configurations, and state changes within software infrastructures. These logs serve as the primary data source for evaluating system health, troubleshooting malfunctions, and detecting security risks. Log Anomaly Detection [16, 17] is a critical task of autonomously identifying patterns in extensive data streams that diverge from "normal" system behaviour. Efficient log anomaly detection enables proactive measures, expediting root cause analysis and ensuring the dependability and availability of essential services.

Modern IT environment is characterized by its distributed architecture, utilizing a variety of components such as cloud services, hybrid-cloud configurations, and proprietary software (e.g., Apache Hadoop, Linux kernels, HDFS). This heterogeneity requires Cross-Domain Log Anomaly Detection, which is the capability to transfer anomaly detection knowledge acquired from one system (the source domain) to a less-monitored system (the target domain). The objective is to transfer detection capabilities across systems that, although functionally linked, exhibit differing log formats, syntax, and distributions of operational events. This method is driven by the necessity to safeguard and oversee new environments without protracted, resource-intensive task of manually collecting adequate labeled anomaly data specific to that system.

The following are the critical challenges involved in the log anomaly detection:

1. ***Extreme Class Imbalance [18]*:** It is observed to having severe disparity between normal and anomaly events.
2. ***Log Heterogeneity and Data Drift [19]*:** Logs are inherently unstructured or semi-structured, varying wildly in format and semantics across different sources.
3. ***Label Scarcity Problem [16]*:** In a new target domain, there is a fundamental lack of labelled anomaly data rely heavily on large, accurately labelled datasets.
4. ***Feature Extraction Complexity*:** To detect anomalies, logs must be converted from raw text into meaningful numerical feature vectors. This requires the mechanisms to capture both the semantic and structural sequence of events.

To address the challenges, in this paper, we proposed model agnostic meta learning and prototypical networks for log anomaly detection problem. The extremeness int the imbalance is handled by SMOTE and feature extraction by BERT and feature selection techniques are also employed to reduce the time complexity.

The paper is organized as follows: Section 2 provides the literature review; Section 3 presents the proposed methodology; results are analysed in Section 4 and conclusions are provided in Section 5.

## 2    Related Work

Le et al. [1] proposed NeuralLog wherein the first stage, the semantic embeddings are extracted over the received raw log messages. They employed transformer architecture to detect the anomalies. Qi et al. [2] employed ChatGPT for identifying the anomalies in the provided logs. Their results demonstrated that the ChatGPT based detection outperformed three deep learning based methods on BGL and spirit datasets. Chen et al. [3] proposed BERT-log where the logs are treated as the natural language sequences and BERT model is fine-tuned to detect the anomalies. Wang et al. [4] proposed Lightlog which involves the generation of low dimensional semantic vectors by utilizing word2vec and post-processing algorithm (PPA). They employed lighweight temporal convolutional network as the classifier. Guo et al. [5] proposed LogBERT, where they employed self-supervised learning mechanism along with the bi-directional LSTM (Bi-LSTM) network for the log anomaly detection. Yang et al. [6] proposed PLElog, semi-supervised approach to label the unlabeled data. This method considers the knowledge obtained from the historical patterns via probabilis-tic label estimation. Almodovar et al. [7] proposed LogFiT model, is also an unsupervised learning approach. This method employs Bi-LSTM to understand the anomaly pattern and then label the data. Li et al. [8] proposed SwissLog, which constructs relational graphs across the distributed components and groups the messages through their respective IDs. This method works as online parser thereby requiring no parameters to tune. This method employed Bi-LSTM model to discriminate the anomaly pattern from the normal ones. Lee et al. [9] proposed LAnoBERT, which does not utilize any log parser and employed BERT model to extract the features. Their results indicate that LAnoBERT outperformed several methods in HDFS, BGL and Thunderbird methods. Zhang et al. [10] proposed DeepTralog, begins the detection process by generating the graph from the provided logs. They employed graph neural networks based Deep SVDD model where it combines logs and detect the anomalies. Vervaet et al. [11] proposed distributed based approach for log detection and named it Monilog. Jia et al. [12] proposed LogFlash, real-time streaming detection method thereby enabling both training and detection in real-time. Catillo et al. [13] proposed Autolog, which operates in two stage manner, wherein the first stage computes scores for the given score and then auto encoder (AE) is employed to reconstruct the scores. The anomaly behaviour is identified based on the user-defined threshold over the reconstruction error.

## 3     Proposed Methodology

The proposed framework is suitable for cross domain log anomaly detection where the following steps are invoked in a sequential manner:

(i)     **_Log Parsing_**: Logs are collected from varied sources i.e., Apache Hadoop, Linux, HDFS etc. It is important to note that the logs collected from these sources follows its own template. We employed Drain3 parser to extract the structural patterns by considering the source-specific log templates.

(ii)     **_Drift based Labeling:_** Due to collection of logs from heterogenous sources; the patterns of being anomaly is distinct from one source to the other. Hence, we employed semantic and fuzzy matching to transfer anomaly knowledge from extant sources to the new incoming logs. This helps to identify the new anomaly patterns effectively.

(iii)     **_Feature Extraction:_** We employed BERT model to extract the features from the labeled dataset obtained from the previous steps. This resulted in 786 dimensional embeddings. Further Drain3 also gives the structural embeddings which adds another 80 features making them to 848 dimensional embeddings.

(iv)     **_Feature Selection through Pyspark:_** We employed Mutual Information (MI) and random forest based scoring from PySpark to extract the top $K$ optimal number of features. In our settings, we fixed the value of $K$ to be 200. This number was fixed after thorough experimentation.

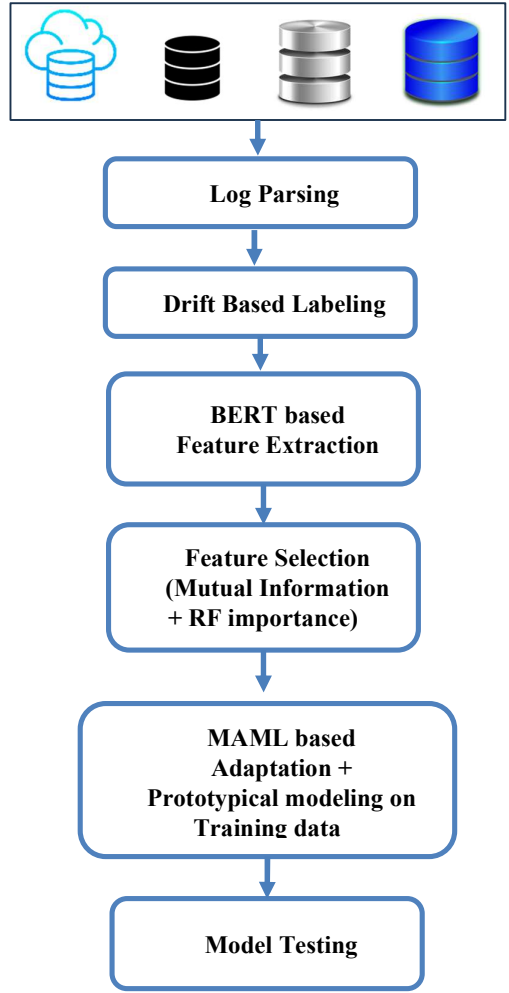

**Fig. 1. Overview of the proposed methodology**

*(i)*    ***Training and Test split*** The dataset is now divided into training and test sets with 70%:30% proportion.

(ii)   ***Invoking Balancing techniques:*** We employed Synthetic minority oversampling technique (SMOTE) to balance the training dataset. Further, we combined the balancing technique with Focal loss to handle the severed imbalance in the dataset.

(iii)  ***Training phase (MAML based adaptation):*** In this step, model agnostic meta learning approach is invoked to get the good initial parameter space for the model. We employed prototypical networks that calculates the distance of the mean vectors to discriminate the normal and anomaly patterns.

Further, this model is trained using a 3-phase curriculum where each task requires only 5-minority examples.

*(iv)* ***Testing phase:*** The effectiveness of the trained model obtained from the previous is now tested over the test dataset and the metrics are reported thereof.

## 4 Results & Discussion

In this section, we first discuss the dataset description followed by the experimental setup and later analyzes the performance of the proposed approach.

### 4.1 Dataset Description

The logs are collected from heterogenous sources and the details of the datasets are outlined in Table 1. These logs are collected from the loghub github repository [14]. Overall, the dataset is collected from 16 varied unique resources and each sources containing with distinct number of logs. All these log datasets are highly imbalance in nature and we also included imbalance ratio present in the dataset individually and the overall is also mentioned in Table 1. In summary, 32,000+ logs are obtained from 16 unique sources.

**Table 1. Details of the datasets collected from heterogenous sources**

| Source Category | Total Logs | Imbalance Ratio |
|---|---|---|
| Web Servers (Apache) | 2,000 | 3.19:1 |
| Operating Systems (Linux, Mac) | 4,000 | 8.5:1 |
| Mobile (Android) | 2,000 | 73.07:1 |
| Healthcare (HealthApp) | 2,000 | 332.33:1 |
| Big Data (Hadoop, Spark, HDFS) | 6,000 | 145.2:1 |
| Security (OpenSSH, OpenStack) | 4,000 | 12.8:1 |
| Other (BGL, Thunderbird, Zookeeper, HPC, Proxifier, Windows) | 12,000 | 42.6:1 |
| **Total** | **32,000** | **62.5:1 (overall)** |

### 4.2 Experimental Setup

We implemented the entire project with Pyspark 3.4 in google colab under pseudo-distributed settings. We utilized Python 3.11 and Pytorch 2.0 with CUDA 18 to develop the deep learning based models. All the pretrained models are obtained from hugging face library. Further, we deployed our proposed model by using Hugging Face Hub [15].

### 4.3 Evaluation Metrics

F1 Score (see Eq. 1) is defined as the harmonic mean of precision and recall. This metric is specifically employed since it offers a balances the model's ability to correctly identify positive cases (Recall) against the accuracy of its positive predictions (Precision).

$$F1score = \frac{2 * (Precision + Recall)}{Precision * Recall} \qquad (1)$$

### 4.4 Discussion over results

All the models are trained by utilizing leave one source out (LOSO) cross validation strategy. Since the number of sources are 16, by using this method, we trained models over 15 sources and tests the logs obtained from the held-out source. The results obtained by the proposed method is presented in Table 2.

The results of Mean F1 score and the corresponding standard deviation is presented in the Table 2. We have specifically chosen F1 score as the metric since this metric is robust to the highly imbalanced in nature. Among the models, our proposed Meta-learning approach turned out to be the best and secured Rank 1 by obtaining mean F1 score of 94.2%. Further, this higher mean F1 score is accompanied by a relatively very less deviation (approximately 6%) across the datasets. This demonstrates the ability of the proposed model being effective and offering stable solution across various datasets. This superior performance is attributed majorly due to the principle of meta learning model that focuses on "learning to learn". Additionally, it also utilizes experiences yielded from multiple sources. This makes it exceptionally well-suited to both quickly adapt to the novel anomaly pattern and yield better generalization across varied data sources.

XGBoost stands second in the list by yielding mean F1 score of 83.8% which is substantially less than that of proposed meta learning approach. Further, XGBoost also exhibits a higher degree of standard deviation ( approximately 13.0%), thereby suggesting the performance of model is less consistent and turned out to be more sensitive to the specific training data. The next-best performers yielded mean F1 score of 50% to 67%, by the remaining deep learning and transformer models indicating poor overall performance. Despite obtaining 67%, the CNN-Attention model shows the worst stability with an extremely high standard deviation of 30.1%. This makes the model highly unreliable choice in a real-world application.

Several models based on the Transformer architecture, such as DeBERTa-v3 and LogBERT are employed as baselines. However, all these models yielded very poor F1 scores, hovering around the 50-52%$ mark. This suggests the models were either poorly adapted to the non-textual or complex structure of the dataset, or that the task's complexity was simply too high for their standard configurations. While their low performance was relatively consistent (approximately 12-15%$), their low F1 score still relegates them to the bottom tier of the ranking. The overall results strongly favour the Meta-Learning paradigm for this particular problem, suggesting a need for more adaptive and specialized learning strategies over generalized deep learning models.

**Table 2. Results obtained by various models**

| Rank | Model | Mean F1 score (+/- Std) |
|------|-------|-------------------------|
| 1 | **Meta-Learning** | **94.2% +/- 6.0%** |
| 2 | XGBoost | 83.8% +/- 13.0% |
| 3 | CNN-Attention | 67.0% +/- 30.1% |
| 4 | Stacked AE | 55.2% +/- 23.1% |
| 5 | DeBERTa-v3 | 52.2% +/- 15.5% |
| 6 | FLNN | 52.3% +/- 22.8% |
| 7 | TabNet | 52.1% +/- 21.6% |
| 8 | LogBERT | 51.1% +/- 12.1 % |
| 9 | VAE | 50.9% +/- 20.3% |
| 10 | DAPT BERT | 50.2 % +/- 15.4% |

## 5 Conclusions and Future Work

In this paper, we proposed meta learning driven cross-domain log anomaly detection method which effectively handles the extreme class imbalance. Our proposed framework equips with the drift-based labeling and feature selection methods which makes them more effective while handling rare anomaly patterns. Our results indicated that the proposed framework turned out to be the most effective in terms of mean F1 score outperforming the state-of-the-art methods.

In future, we will focus on the model that works under low resource constraints by adapting model distillation kind of techniques.

## References

1. Le, V. H., & Zhang, H. (2021, November). Log-based anomaly detection without log parsing. In *2021 36th IEEE/ACM International Conference on Automated Software Engineering (ASE)* (pp. 492-504). IEEE.
2. Qi, J., Huang, S., Luan, Z., Yang, S., Fung, C., Yang, H., ... & Wu, Z. (2023, December). Loggpt: Exploring chatgpt for log-based anomaly detection. In *2023 IEEE International Conference on High Performance Computing & Communications, Data Science & Systems, Smart City & Dependability in Sensor, Cloud & Big Data Systems & Application (HPCC/DSS/SmartCity/DependSys)* (pp. 273-280). IEEE.
3. Chen, S., & Liao, H. (2022). Bert-log: Anomaly detection for system logs based on pre-trained language model. *Applied Artificial Intelligence*, *36*(1), 2145642.
4. Wang, Z., Tian, J., Fang, H., Chen, L., & Qin, J. (2022). LightLog: A lightweight temporal convolutional network for log anomaly detection on the edge. *Computer Networks*, *203*, 108616.
5. Guo, H., Yuan, S., & Wu, X. (2021, July). Logbert: Log anomaly detection via bert. In *2021 international joint conference on neural networks (IJCNN)* (pp. 1-8). IEEE.

6. Yang, L., Chen, J., Wang, Z., Wang, W., Jiang, J., Dong, X., & Zhang, W. (2021, May). Semi-supervised log-based anomaly detection via probabilistic label estimation. In *2021 IEEE/ACM 43rd International Conference on Software Engineering (ICSE)* (pp. 1448-1460). IEEE.

7. Almodovar, C., Sabrina, F., Karimi, S., & Azad, S. (2024). LogFiT: Log anomaly detection using fine-tuned language models. *IEEE Transactions on Network and Service Management*, *21*(2), 1715-1723.

8. Li, X., Chen, P., Jing, L., He, Z., & Yu, G. (2022). SwissLog: Robust anomaly detection and localization for interleaved unstructured logs. *IEEE Transactions on Dependable and Secure Computing*, *20*(4), 2762-2780.

9. Lee, Y., Kim, J., & Kang, P. (2023). Lanobert: System log anomaly detection based on bert masked language model. *Applied Soft Computing*, *146*, 110689.

10. Zhang, C., Peng, X., Sha, C., Zhang, K., Fu, Z., Wu, X., ... & Zhang, D. (2022, May). Deeptralog: Trace-log combined microservice anomaly detection through graph-based deep learning. In *Proceedings of the 44th international conference on software engineering* (pp. 623-634).

11. Vervaet, A. (2021, April). Monilog: An automated log-based anomaly detection system for cloud computing infrastructures. In *2021 IEEE 37th international conference on data engineering (ICDE)* (pp. 2739-2743). IEEE.

12. Jia, T., Wu, Y., Hou, C., & Li, Y. (2021, October). Logflash: Real-time streaming anomaly detection and diagnosis from system logs for large-scale software systems. In *2021 IEEE 32nd International Symposium on Software Reliability Engineering (ISSRE)* (pp. 80-90). IEEE.

13. Catillo, M., Pecchia, A., & Villano, U. (2022). AutoLog: Anomaly detection by deep autoencoding of system logs. *Expert Systems with Applications*, *191*, 116263.

14. Loghub github repository: https://github.com/logpai/loghub/tree/master

15. Hugging Face: https://huggingface.co/

16. Meena Siwach, D. S. M. (2022). Anomaly detection for web log data analysis: A review. *Journal of Algebraic Statistics*, *13*(1), 129-148.

17. Le, V. H., & Zhang, H. (2022, May). Log-based anomaly detection with deep learning: How far are we?. In *Proceedings of the 44th international conference on software engineering* (pp. 1356-1367).

18. Ma, X., Zou, H., He, P., Keung, J., Li, Y., Yu, X., & Sarro, F. (2024). On the influence of data resampling for deep learning-based log anomaly detection: Insights and recommendations. *IEEE Transactions on Software Engineering*.

19. Wani, D., Ackerman, S., Farchi, E., Liu, X., Chang, H. W., & Lalithsena, S. (2023). Data Drift Monitoring for Log Anomaly Detection Pipelines. *arXiv preprint arXiv:2310.14893*.